\documentclass[runningheads]{llncs}
\usepackage{graphicx}
\usepackage{amsmath,amssymb} 
\usepackage{color}

\usepackage{hyperref}
\usepackage{pifont}
\usepackage{multirow}
\usepackage{array}
\usepackage{subfig}
\usepackage{hyphenat}
\usepackage{float}

\def \etal {{\em et~al.}}
\def\eg{{\em e.g.}}

\newcommand{\cmark}{\ding{51}}
\newcommand{\xmark}{\ding{55}}

\newcommand{\PreserveBackslash}[1]{\let\temp=\\#1\let\\=\temp}
\newcolumntype{C}[1]{>{\PreserveBackslash\centering}p{#1}}

\newcommand{\mb}[1]{\mathbf{#1}}
\newcommand{\mr}[1]{\mathrm{#1}}
\newcommand{\mc}[1]{\mathcal{#1}}
\newcommand{\norm}[1]{\left\lVert #1 \right\rVert_2}

\newcommand{\Rn}[1]{\uppercase\expandafter{\romannumeral#1}}

\DeclareMathOperator*{\argmax}{arg\,max}

\begin{document}
\pagestyle{headings}
\mainmatter

\def\ACCV18SubNumber{29}  

\title{Semi-Supervised Learning for Face Sketch Synthesis in the Wild} 
\titlerunning{Face Sketch in the Wild}
\authorrunning{Chen \etal}

\author{Chaofeng Chen$^1$, Wei Liu$^1$, Xiao Tan$^2$ and Kwan-Yee K. Wong$^1$}
\institute{$^1$The University of Hong Kong, $^2$Baidu Research
\\{\tt\small \{cfchen, wliu, kykwong\}@cs.hku.hk, tanxchong@gmail.com}}
\maketitle

\begin{abstract}
Face sketch synthesis has made great progress in the past few years. Recent methods based on deep neural networks are able to generate high quality sketches from face photos. However, due to the lack of training data (photo-sketch pairs), none of such deep learning based methods can be applied successfully to face photos in the wild. In this paper, we propose a semi-supervised deep learning architecture which extends face sketch synthesis to handle face photos in the wild by exploiting additional face photos in training. Instead of supervising the network with ground truth sketches, we first perform patch matching in feature space between the input photo and photos in a small reference set of photo-sketch pairs. We then compose a {\em pseudo sketch feature} representation using the corresponding sketch feature patches to supervise our network. With the proposed approach, we can train our networks using a small reference set of photo-sketch pairs together with a large face photo dataset without ground truth sketches. Experiments show that our method achieves state-of-the-art performance both on public benchmarks and face photos in the wild. Codes are available at \url{https://github.com/chaofengc/Face-Sketch-Wild}.
\end{abstract}

\section{Introduction}\label{sec:intro}
Face sketch synthesis targets at generating a sketch from an input face photo. It has many useful applications. For instance, police officers often have to rely on face sketches to identify suspects, and face sketch synthesis makes it feasible for matching sketches against photos in a mugshot database automatically. Artists can also employ face sketch synthesis to simplify the animation production process \cite{song2014real}. Many people prefer using sketches as their profile pictures in social media networks \cite{berger2013style}, and face sketch synthesis allows them to produce sketches without the help of a professional artist.

Much effort has been devoted to face sketch synthesis. In particular, exemplar based methods dominated in the past two decades. These methods can achieve good performance without explicitly modeling the highly nonlinear mapping between face photos and sketches. They commonly subdivide a test photo into overlapping patches, and match these test patches with the photo patches in a reference set of photo-sketch pairs. They then compose an output sketch using the corresponding sketch patches in the reference set. Although promising results have been reported \cite{song2014real,zhou2012markov,ijcai2017-500,wang2009face}, these methods have several drawbacks. For example, sketches in Fig.~\ref{fig:result_bmk}(c)(d)(e)(f) are over-smoothed and fail to preserve subtle contents such as strands of hair on the forehead. Moreover, the patch matching and optimization processes are often very time-consuming. Recent methods exploited Convolutional Neural Networks (CNNs) to learn a direct mapping between photos and sketches, which is, however, a non-trivial task. The straight forward CNN based method produces blurry sketches (see Fig.~\ref{fig:result_bmk}(g)), and methods based on Generative Adversary Networks (GAN) \cite{goodfellow2014generative} introduce undesirable artifacts (see Fig.~\ref{fig:result_bmk}(h),(i)). Besides, all these CNN based methods do not generalize well to face photos in the wild due to the lack of large training datasets of photo-sketch pairs. Although unpaired GAN based methods such as Cycle-GAN \cite{CycleGAN2017} can use unpaired data to transfer images between different domains, they fail to well preserve the facial content because of the weak content constraint (see fig.~\ref{fig:result-wild}).

In this paper, we propose a semi-supervised learning framework for face sketch synthesis that takes advantages of the exemplar based approach, the perceptual loss and GAN. We design a residual net \cite{He2015} with skip connections as our generator network. Suppose we have a small reference set of photo-sketch pairs and a large face photo dataset without ground truth sketches. Similar to the exemplar based approach, we subdivide the VGG-19 \cite{simonyan2014very} feature maps of the input photo into overlapping patches, and match them with the photo patches (in feature space) in the reference set. We then compose a {\em pseudo sketch feature} representation using the corresponding sketch patches (in feature space) in the reference set.  We can then supervise our generator network using a perceptual loss based on the mean squared error (MSE) between the feature maps of the generated sketch and the corresponding pseudo sketch feature of the input photo. An adversary loss is also utilized to make the generated sketches more realistic.

In summary, our main contributions are three folds: (1) A semi-supervised learning framework for face sketch synthesis. Our framework allows us to train our networks using a small reference set of photo-sketch pairs together with a large face photo dataset without ground truth sketches. This enables our networks to generalize well to face photos in the wild. (2) A perceptual loss based on pseudo sketch feature. We show that the proposed loss is critical in preserving both facial content and texture details in the generated sketches. Extensive experiments are conducted to verify the effectiveness of our model. Both qualitative and quantitative results illustrate the superiority of our method. (3) To the best of our knowledge, our method is the first work that can generate visually pleasant sketches for face photos in the wild.

\section{Related Works}

\subsection{Exemplar Based Methods}

Tang and Wang \cite{tang2003face} first introduced the exemplar based method based on eigentransformation. They projected an input photo onto the eignspace of the training photos, and then reconstruct a sketch from the eignspace of the training sketches using the same projection. Liu~\etal~\cite{liu2005nonlinear} observed that the linear model holds better locally, and therefore proposed a nonlinear model based on local linear embedding (LLE). They first subdivided an input photo into overlapping patches and reconstructed each photo patch as a linear combination of the training photo patches. They then obtained the sketch patches by applying the same linear combinations to the corresponding training sketch patches. Wang and Tang \cite{wang2009face} employed a multi-scale markov random fields (MRF) model to improve the consistency between neighboring patches. By introducing shape priors and SIFT features, Zhang~\etal~\cite{zhang2010lighting} proposed an extended version of MRF which can handle face photos under different illuminations and poses. However, these MRF based methods are not capable of synthesizing new sketch patches since they only select the best candidate sketch patch for each photo patch. To tackle this problem, Zhou~\etal~\cite{zhou2012markov} presented the markov weight fields (MWF) model which produces a target sketch patch as a linear combination of $K$  best candidate sketch patches. Considering that patch matching based on traditional image features (e.g., PCA and SIFT) is not robust, a recent method \cite{ijcai2017-500} used CNN feature to represent the training patches and computed more accurate combination coefficients. To accelerate the synthesis procedure, Song~\etal~\cite{song2014real} formulated face sketch synthesis as a spatial sketch denoising (SSD) problem, and Wang~\etal~\cite{wangrslcr} presented an offline random sampling strategy for nearest neighbor selection of patches.

\subsection{Learning Based Methods}

Recent works applied CNN to synthesize sketches and produced promising results. Zhang~\etal~\cite{zhang2015end} proposed a 7-layer fully convolutional network (FCN) to directly transfer an input photo to a sketch. Although their model can roughly estimate the outline of a face, it fails to capture texture details with the use of intensity based mean squared error (MSE) loss. 
Zhang~\etal~\cite{zhang2017content} utilized a branched fully convolutional network (BFCN) consisting of a content branch and a texture branch. Because the face content and texture are predicted separately with different loss metrics, the final sketch looks disunited. Chen~\etal~\cite{chen2017pcf} proposed the pyramid column feature and used it to compose a reference style for a test photo from the training sketches. They utilized a CNN to create a content image from the photo, and then transferred the reference style to introduce shadings and textures in the output sketch. Wang~\etal~\cite{wang2017data} presented the multi-scale generative adversarial networks (GANs) to generate sketches from photos and vice versa. Multiple discriminators at different hidden layers are used to supervise the synthesis process. Gao~\etal~\cite{Gao2017cagan} took advantage of the facial parsing map and proposed a composition-aided stack GAN. All these deep learning based methods require ground truth photo-sketch pairs for training, and they do not generalize well to face photos in the wild due to the lack of training data.

\section{Semi-Supervised Face Sketch Synthesis} \label{sec:semi}

\subsection{Overview}

\begin{figure*}[htbp]
\begin{center}
   \includegraphics[width=0.99\linewidth]{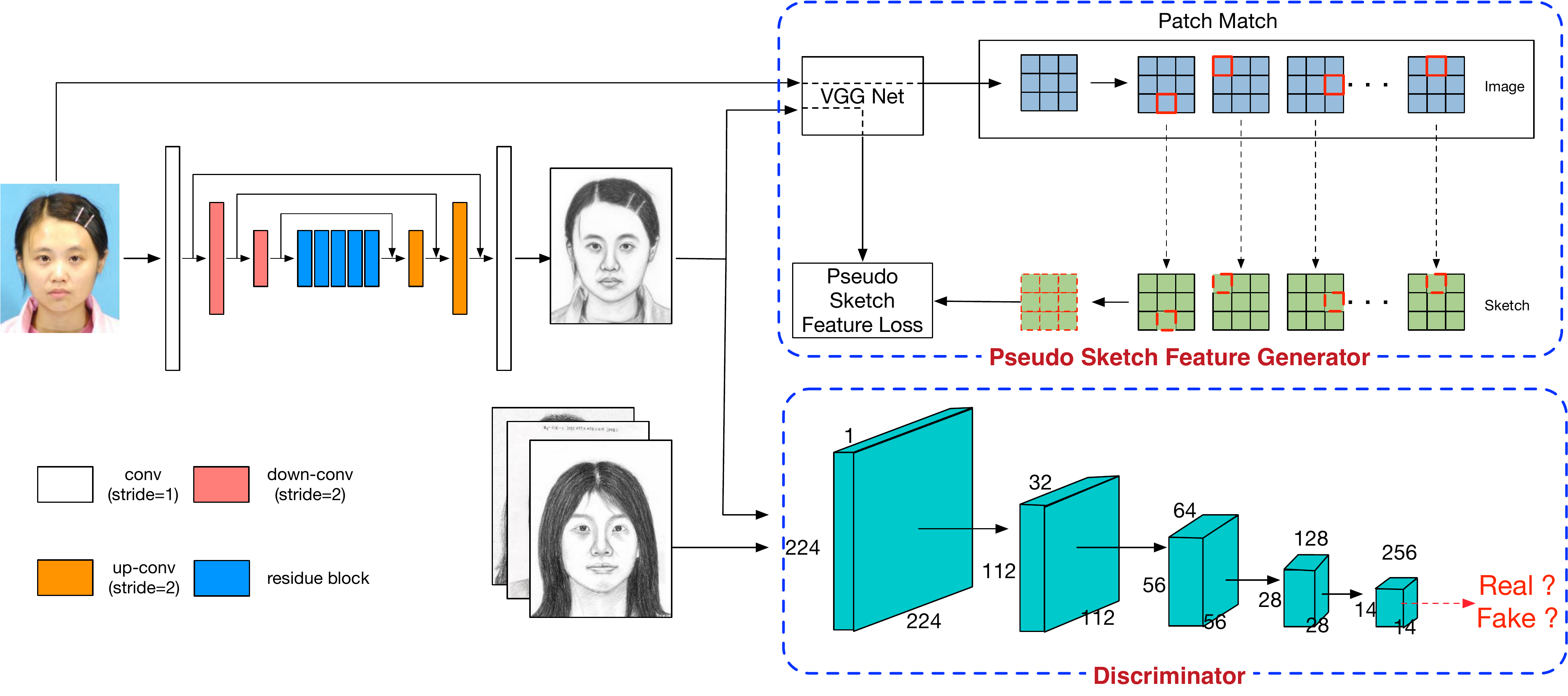}
\end{center}
   \caption{Framework of the proposed method. The generator network is a deep residual network with skip connections. It generates a synthesized sketch from an input photo. The pseudo sketch feature generator utilizes patch matching in the deep feature space to generate a pseudo sketch feature for an input photo in training. The discriminator network tries to distinguish between generated sketches and sketches drawn by artists.}
\label{fig:architecture}
\end{figure*}

Our framework is composed of three main parts, namely a generator network $G$, a pseudo sketch feature generator and a discriminator network $D$ (see Fig.~\ref{fig:architecture}). The generator network is a deep residual network with skip connections. It is used to generate a synthesized sketch $\hat{\mb{y}}$ for each input photo $\mb{x}$. The pseudo sketch feature generator is the key to our semi-supervised learning approach. Instead of training the generator network directly with ground truth sketches, we construct a pseudo sketch feature for each input photo to supervise the synthesis of $\hat{\mb{y}}$. In this way, we can train our network on any face photo datasets, and generalize our model to face photos in the wild. We further adopt a discriminator network $D$ to minimize the gap between generated sketches and real sketches drawn by artists.

\subsection{Pseudo Sketch Feature Generator}
\begin{figure*}
  \centering
    \subfloat[]{\includegraphics[width=0.33\linewidth]{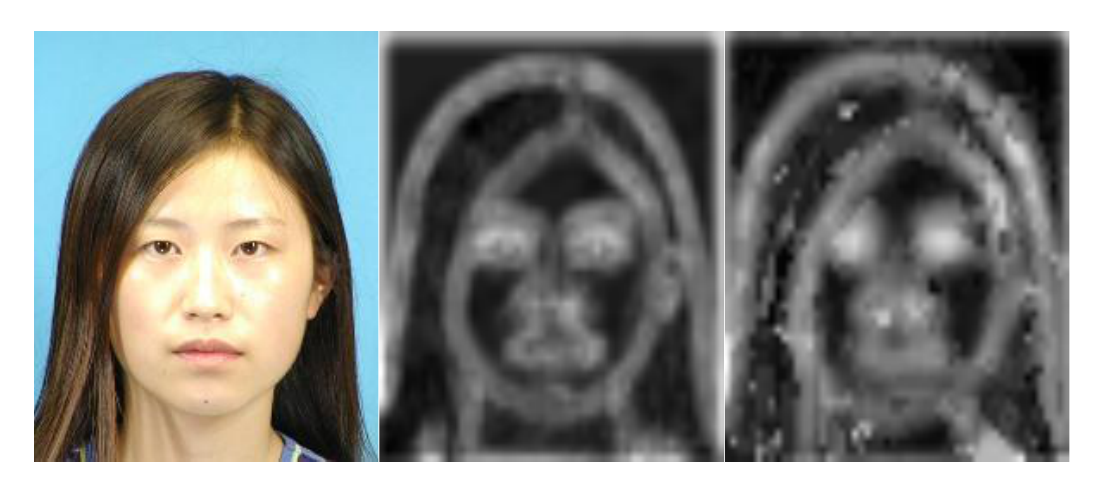}}
    \subfloat[]{\includegraphics[width=0.24\linewidth]{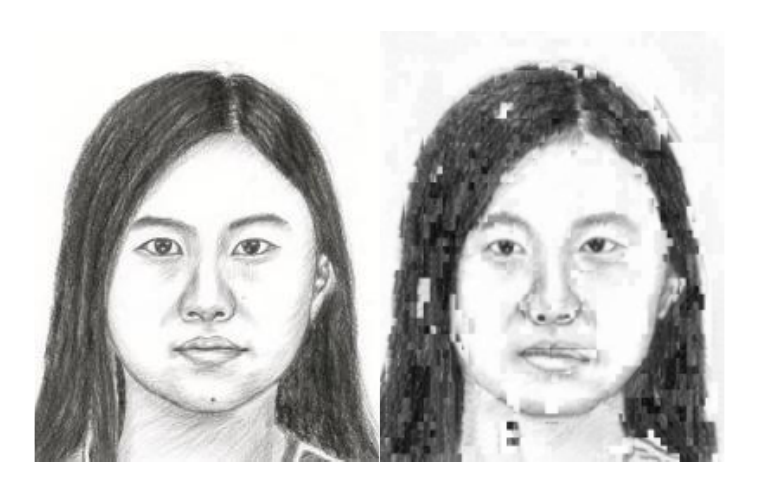}}
    \subfloat[]{\includegraphics[width=0.33\linewidth]{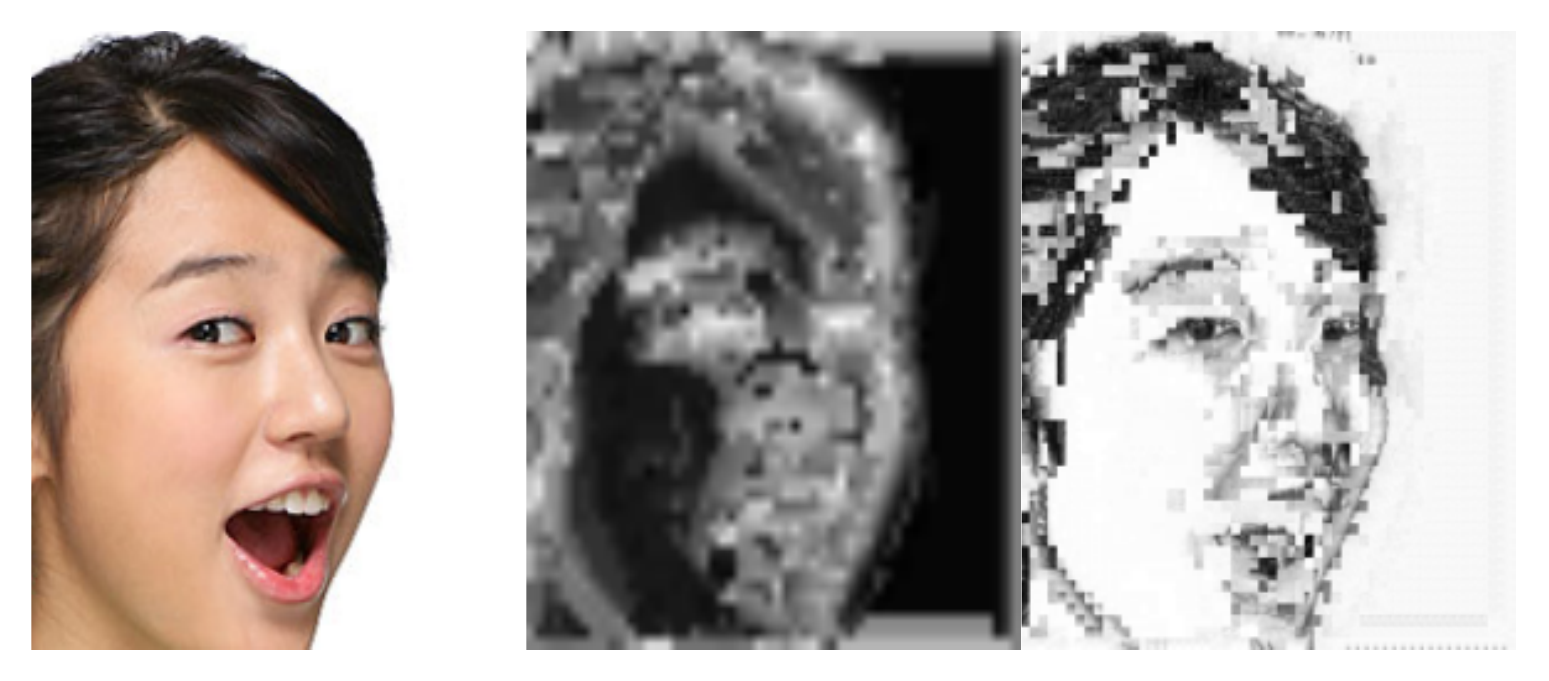}}
    \caption{(a) Ground truth sketch feature (middle) and pseudo sketch feature of the relu3\_1 layer (right). (b) Ground truth sketch (left) and pixel level projection of the patch matching result (right). (c) Photos in the wild without ground truth sketches. (\textit{Note that the pixel level results are only for visualization, and they are not used in training.})}
    \label{fig:pm-example}
\end{figure*}

Given a reference set $\mc{R} = \{(\mb{x}^\mc{R}_i, \mb{y}^\mc{R}_i)\}_{i=1}^N$, the pseudo sketch feature generator targets at constructing a pseudo sketch feature $\mr{\Phi}'({\mb{x}})$ for a test photo $\mb{x}$ which is used to supervise the synthesis of the sketch $\hat{\mb{y}}$. We follow MRF-CNN \cite{li2016combining} to extract a local patch representation of an image.
We first feed $\mb{x}$ into a pretrained VGG-19 network and extract the feature map $\mr{\Phi}^l(\mb{x})$ at the $l$-th layer. Similarly, we obtain $\{\mr{\Phi}^l(\mb{x}^\mc{R}_i)\}_{i=1}^N$ and $\{\mr{\Phi}^l(\mb{y}^\mc{R}_i)\}_{i=1}^N$. Let us denote a $k \times k$ patch centered at a point $j$ of $\mr{\Phi}^l(\mb{x})$ as $\mr{\Psi}_j\left( \mr{\Phi}^l(\mb{x})\right)$, and the same definition applies to $\mr{\Psi}_j\left( \mr{\Phi}^l(\mb{x}^\mc{R}_i)\right)$ and $\mr{\Psi}_j(\mr{\Phi}^l\left( \mb{y}^\mc{R}_i)\right)$. Now for each patch $\mr{\Psi}_j\left( \mr{\Phi}^l(\mb{x})\right)$, where $j = 1, 2, \ldots, m$ and $m = (H^l - 2\times \lfloor \frac{k}{2} \rfloor)\times(W^l - 2 \times \lfloor \frac{k}{2} \rfloor)$ with $H^l$ and $W^l$ being the height and width of $\mr{\Phi}^l(\mb{x})$, we find its best match $\mr{\Psi}_{j'}\left( \mr{\Phi}^l(\mb{x}^\mc{R}_{i'})\right)$ in the reference set based on cosine distance, i.e.,
\begin{equation}
  (i', j') = \argmax_{\begin{subarray}{c} i^*=1\sim N \\ j^*=1\sim m \end{subarray}} \frac{\mr{\Psi}_j\left( \mr{\Phi}^l(\mb{x})\right) \cdot \mr{\Psi}_{j^*}\left( \mr{\Phi}^l(\mb{x}_{i^*}^\mc{R})\right)}{\norm{\mr{\Psi}_j\left( \mr{\Phi}^l(\mb{x})\right)} \norm{\mr{\Psi}_{j^*}\left( \mr{\Phi}^l(\mb{x}_{i^*}^\mc{R})\right)}}.
  \label{equ:cosine_dist}
\end{equation}
Since the photos and the corresponding sketches in $\mc{R}$ are well aligned, we directly apply $(i', j')$ to index the corresponding sketch feature patch $\mr{\Psi}_{j'}\left( \mr{\Phi}^l(\mb{y}^\mc{R}_{i'})\right)$ for $\mr{\Psi}_{j'}\left( \mr{\Phi}^l(\mb{x}^\mc{R}_{i'})\right)$, and use it as the pseudo sketch feature patch $\mr{\Psi}'_j\left( \mr{\Phi}^l(\mb{x})\right)$ for $\mr{\Psi}_j\left( \mr{\Phi}^l(\mb{x})\right)$. Finally, a pseudo sketch feature representation (at layer $l$) for $\mb{x}$  is given by $\{\mr{\Psi}'_j\left( \mr{\Phi}^l(\mb{x})\right)\}_{j=1}^m$. Fig. \ref{fig:pm-example} visualizes an example of the pseudo sketch feature. It can be seen that the pseudo sketch feature provides a good approximation of the real sketch feature (see Fig.~\ref{fig:pm-example}(a)). We also show a na\"{i}ve reconstruction  in Fig.~\ref{fig:pm-example}(b) obtained by directly using the matching index to index the pixel values in the training sketches. We can see such a na\"{i}ve reconstruction does roughly resemble the real sketch, which also justifies the effectiveness of the pseudo sketch feature. Note that we only need alignment between photos and sketches in $\mc{R}$. Since we perform a dense patch matching between the input photo and the reference photos, we can also generate reasonable pseudo sketch features for input faces under different poses (see Fig.~\ref{fig:pm-example}(c)).

\subsection{Loss Functions}

\begin{figure*}
  \captionsetup[subfigure]{labelformat=empty}
  \centering
    \subfloat[Photo \& Sketch]{\includegraphics[width=0.12\linewidth]{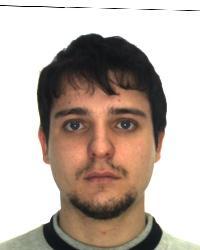}
                \includegraphics[width=0.12\linewidth]{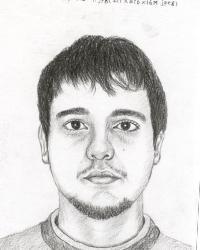}}
    \subfloat[relu1\_1]{\includegraphics[width=0.12\linewidth]{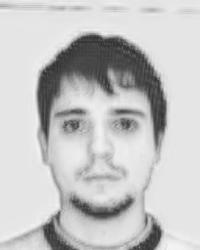}}
    \subfloat[relu2\_1]{\includegraphics[width=0.12\linewidth]{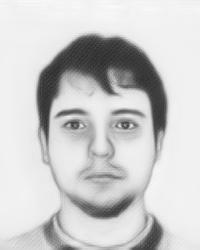}}
    \subfloat[relu3\_1]{\includegraphics[width=0.12\linewidth]{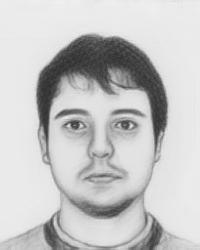}}
    \subfloat[relu4\_1]{\includegraphics[width=0.12\linewidth]{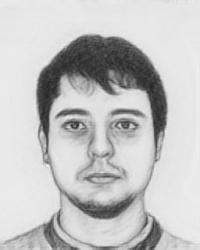}}
    \subfloat[relu5\_1]{\includegraphics[width=0.12\linewidth]{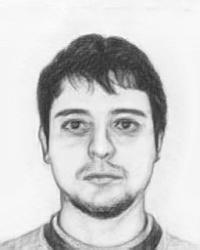}}
    \caption{Results of using different layers in pseudo sketch feature loss.}
    \label{fig:layer-example}
\end{figure*}

\paragraph{Pseudo Sketch Feature Loss}
We define our pseudo sketch feature loss as
\begin{equation}
    L_p(\mb{x}, \hat{\mb{y}}) = \sum_{l=3}^5 \sum_{j=1}^m \norm{\mr{\Psi}_j\left( \mr{\Phi}^l(\hat{\mb{y}})\right) - \mr{\Psi}'_j\left( \mr{\Phi}^l(\mb{x})\right)}^2,
\end{equation}
where $l=3, 4, 5$ refer to layers relu3\_1, relu4\_1, and relu5\_1 respectively. High level features after relu3\_1 are better representations of textures and more robust to appearance changes and geometric transforms \cite{li2016combining}. Fig. \ref{fig:layer-example} shows the results of using different layers in $L_p$. As expected, low level features (\eg, relu1\_1 and relu2\_1) fail to generate sketch textures. While high level features (\eg, relu5\_1) can better preserve textures, they produce artifacts in terms of details (see the eyes of sketches in Fig.~\ref{fig:layer-example}). To get better performance and reduce the computation cost of patch matching, we set $l=3, 4, 5$.

\paragraph{GAN Loss} For easier convergence, we use the least square loss when training the GAN, known as LSGAN \cite{mao2017least}. The objective functions of LSGAN are given by
\begin{gather}
L_{GAN\_D} = \frac{1}{2} \mathbb{E}_{y\sim p_{sketch}(y)}[(D(y) - 1)^2] + \frac{1}{2} \mathbb{E}_{x\sim p_{photo}(x)} [(D(G(x)))^2] \\
L_{GAN\_G} = \mathbb{E}_{x\sim p_{photo}(x)} [(D(G(x)) - 1)^2]
\end{gather}

\paragraph{Total Variation Loss} Sketches generated by CNN may be unnatural and noisy. Following previous works \cite{johnson2016perceptual,li2016combining,Kaur2017}, we adopt the {\em total variation loss} as a natural image prior to further improve the sketch quality, 
\begin{equation}
    L_{tv}(\hat{\mb{y}}) = \sum_{u,v}\left((\hat{\mb{y}}_{u+1,v} - \hat{\mb{y}}_{u,v})^2 + (\hat{\mb{y}}_{u,v+1} - \hat{\mb{y}}_{u,v})^2\right),
\end{equation}
where $\hat{\mb{y}}_{u,v}$ denotes the intensity value at $(u,v)$ of the synthesized sketch $\hat{\mb{y}}$.

Based on the above loss terms, we can train our generator network $G$ and discriminator network $D$ using the following two loss functions respectively:
\begin{gather}
L_G = \lambda_p L_p + \lambda_{adv} L_{GAN\_G} + \lambda_{tv} L_{tv},\\
L_D = L_{GAN\_D}
\end{gather}
where $L_G$ and $L_D$ are minimized alternatively until convergence. $\lambda_p$, $\lambda_{adv}$ and $\lambda_{tv}$ are trade-off weights for each loss term respectively.

\section{Implementation Details}

\subsection{Datasets}

\paragraph{Photo-Sketch Pairs} We use two public datasets: the CUFS dataset(consisting of the CUHK student dataset~\cite{tang2003face}, the AR dataset~\cite{martinez1998r}, and the XM2VTS dataset~\cite{Messer99xm2vtsdb}) and the CUFSF dataset~\cite{zhang2011coupled}, to evaluate our model\footnote{Data comes from \url{http://www.ihitworld.com/RSLCR.html}}. The CUFSF dataset is more challenging than the CUFS dataset because (1) the photos were captured under different lighting conditions and (2) the sketches exhibit strong deformation in shape and cannot be aligned with the photos well. Details of these datasets are summarized in Table \ref{tab:dataset}.

\begingroup
\renewcommand{\arraystretch}{1.}
\begin{table}[htbp]
\footnotesize
\caption{Details of benchmark datasets. Align: whether the sketches are well aligned with photos. Var: whether the photos have lighting variations.} \label{tab:dataset}
\centering
\begin{tabular}{c|c|ccccc}
\hline
\multicolumn{2}{c|}{Dataset} & Total Pairs  & Train & Test & Align  & Var   \\ \hline
\multirow{3}{*}{CUFS} &  CUHK    & 188    & 88    & 100  & \cmark & \xmark  \\ \cline{2-7}
&  AR      & 123    & 80    & 43   & \cmark & \xmark  \\ \cline{2-7}
&  XM2VTS  & 295    & 100   & 195  & \xmark & \xmark \\ \cline{1-7}
\multicolumn{2}{c|}{CUFSF}& 1194    & 250   & 944  & \xmark & \cmark   \\ \hline
\end{tabular}
\end{table}
\endgroup

\paragraph{Face Photos} We use the VGG-Face dataset~\cite{Parkhi15} to evaluate our model on photos in the wild. There are 2,622 persons in this dataset and each person has 1,000 photos. We randomly select 2,000 persons for training and the rest for testing. For each person in the training split, we randomly select $\mc{N}$ photos and named the resulting dataset VGG-Face$\mc{N}$\footnote{The dataset will be made available.}, where $\mc{N} = 01, 02, \ldots, 10$. We also randomly select 2 photos for each person in the testing split to construct a VGG test set of 1,244 photos.

\paragraph{Preprocessing} For photos/sketches which have already been aligned and have a size of $250 \times 200$, we leave them unchanged. For the rest, we first detect 68 face landmarks on the image using \texttt{dlib}\footnote{\url{http://dlib.net/}}, and calculate a similarity transform to warp the image into one with the two eyes located at $(75, 125)$ and $(125, 125)$ respectively. We then crop the resulting image to a size of $250 \times 200 $. We simply drop those photos/sketches from which we fail to detect face landmarks.

\subsection{Patch Matching}
As in exemplar based methods, patch matching is a time-consuming process. We accelerate this process in three ways. First, we precompute and store the feature patches for the photos and sketches in the reference set (i.e., $\{\mr{\Psi}_j\left( \mr{\Phi}^l(\mb{x}^\mc{R}_i)\right)\}$ and $\{\mr{\Psi}_j(\mr{\Phi}^l\left( \mb{y}^\mc{R}_i)\right)\}$). Second, instead of searching the whole reference feature set, we first identify $k$ best matched reference photos for each input photo based on the cosine distance of their relu5\_1 feature maps. Patch matching is then restricted within these $k$ reference photos (we set $k=5$ in the whole training process). Third, Equ. \ref{equ:cosine_dist} is implemented as a convolution operator which can be computed efficiently on GPU.

\subsection{Training Details}

We updated the generator and discriminator alternatively at every iteration. The trade-off weights $\lambda_p$, $\lambda_{adv}$  were set to $1$ and $10^3$, and $\lambda_{tv}$ was set to $10^{-5}$ when using CUFS as reference set and $10^{-2}$ when using CUFSF. We implemented our model using \texttt{PyTorch}\footnote{\url{http://pytorch.org/}}, and trained it on a Nvidia Titan X GPU. We used Adam \cite{kingma2014adam} with learning rates from $10^{-3}$ to $10^{-5}$, decreasing with a factor of $10^{-1}$. Data augmentation was done online in the color space (brightness, contrast, saturation and sharpness). Each iteration took about 2s with a batch size of 6, and the model converged after about 5 hours of training.

\section{Evaluation on Public Benchmarks} \label{sec:benchmark}
In this section, we evaluate our model using two public benchmarks, namely CUFS and CUFSF, which were captured under laboratory conditions. We use the training photos from CUFS$\cup$CUFSF to train our networks. When evaluating on CUFS, the reference photo-sketch pairs only comes from CUFS, and the same applies to CUFSF. To demonstrate the effectiveness of our model, we compare our results both qualitatively and quantitatively with seven other methods, namely MWF~\cite{zhou2012markov}, SSD~\cite{song2014real}, RSLCR~\cite{wangrslcr}, DGFL~\cite{ijcai2017-500}, FCN~\cite{zhang2015end}, Pix2Pix-GAN~\cite{pix2pix2017}, and Cycle-GAN~\cite{CycleGAN2017}. We also compare our results quantitatively with the latest GAN based sketch synthesis methods, i.e., PS$^2$-MAN \cite{Wang2017psman} and stack-CA-GAN \cite{Gao2017cagan}. Since the models of their work are not available, we can only compare with the results that are directly taken from their published papers.

\subsection{Qualitative Comparison}
\begin{figure*}[htbp]
\includegraphics[width=0.99\linewidth]{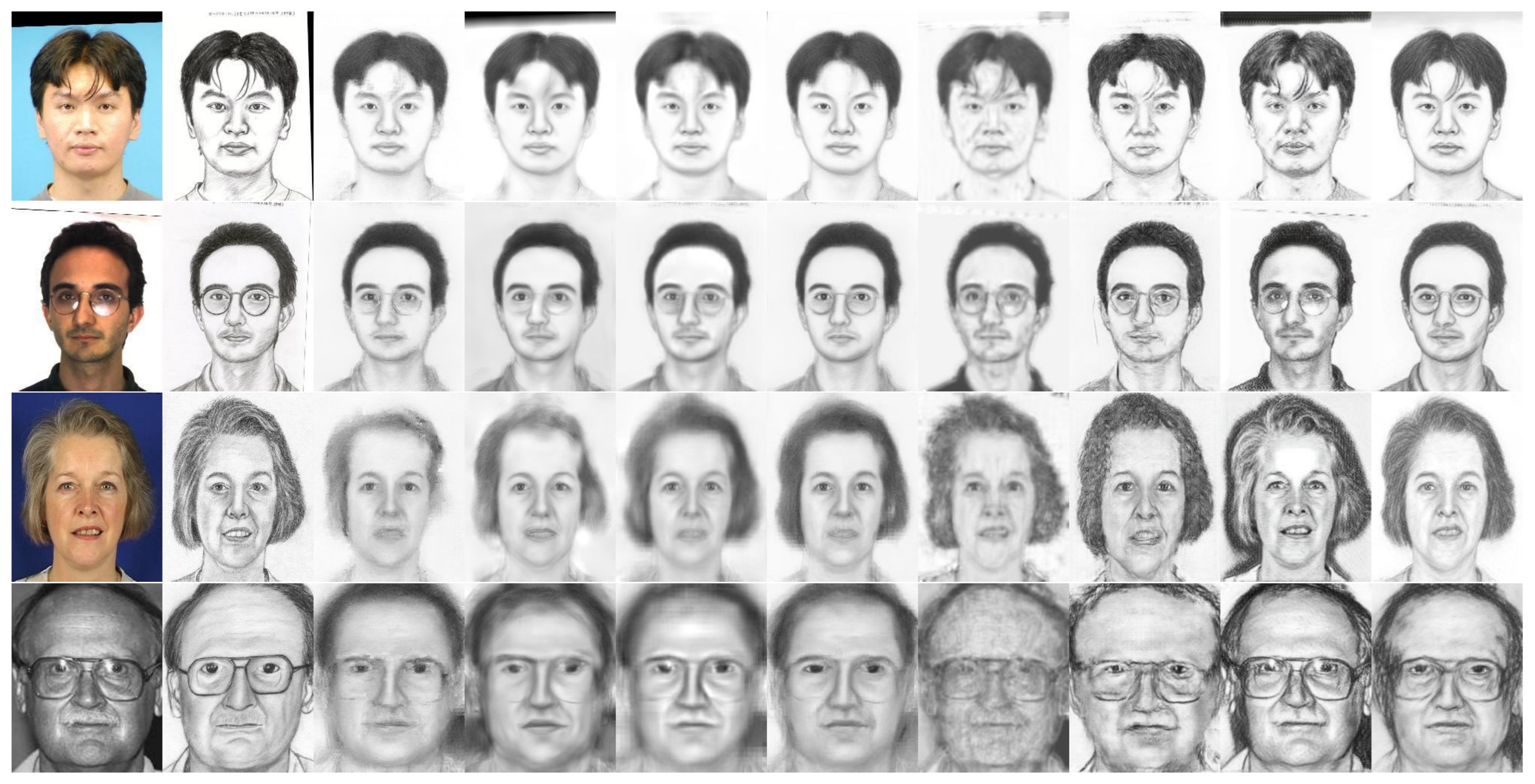}
\begin{minipage}[t]{1.\textwidth}
\begin{tabular}{*{10}{C{1.1cm}}}
(a) & (b) & (c) & (d) & (e) & (f) & (g) & (h) & (i) & (j)\\
Photo & Artist & MWF\cite{zhou2012markov} & SSD\cite{song2014real} & RS-LCR\cite{wangrslcr} & DGFL\cite{ijcai2017-500} & FCN\cite{zhang2015end} & Pix2Pix-GAN\cite{pix2pix2017} & Cycle-GAN\cite{WANG2017} &   Ours
\end{tabular}
\end{minipage}
   \caption{Sketches generated using different methods. First 3 rows: test photos from CUFS. Last row: test photo from CUFSF.}
   \label{fig:result_bmk}
\end{figure*}

As we can observe in Fig.~\ref{fig:result_bmk}, exemplar based methods (see Fig.~\ref{fig:result_bmk}(c),(d),(e) in general perform worse than learning based methods (see Fig.~\ref{fig:result_bmk}(g),(h),(i),(j)), especially in preserving contents of the input photos. Using deep features in exemplar based methods helps to alleviate the problem, but the results are over-smoothed (see Fig.~\ref{fig:result_bmk}(f)). Due to the lack of training data, FCN produces bad results when the photos are taken under very different lighting conditions (see last two rows of Fig.~\ref{fig:result_bmk}(g)). Although the two GANs can produce much better results than FCN, they also introduce many artifacts and noise. Thanks to the pseudo sketch feature loss, our method does not suffer from the above problems. In particular, our semi-supervised strategy allows us to incorporate more training photos without ground truth in training, which helps to improve the generalization ability.

\subsection{Quantitative Comparison}

\subsubsection{Image Quality Assessment}

\begin{figure}[htbp]
\includegraphics[width=1.\linewidth]{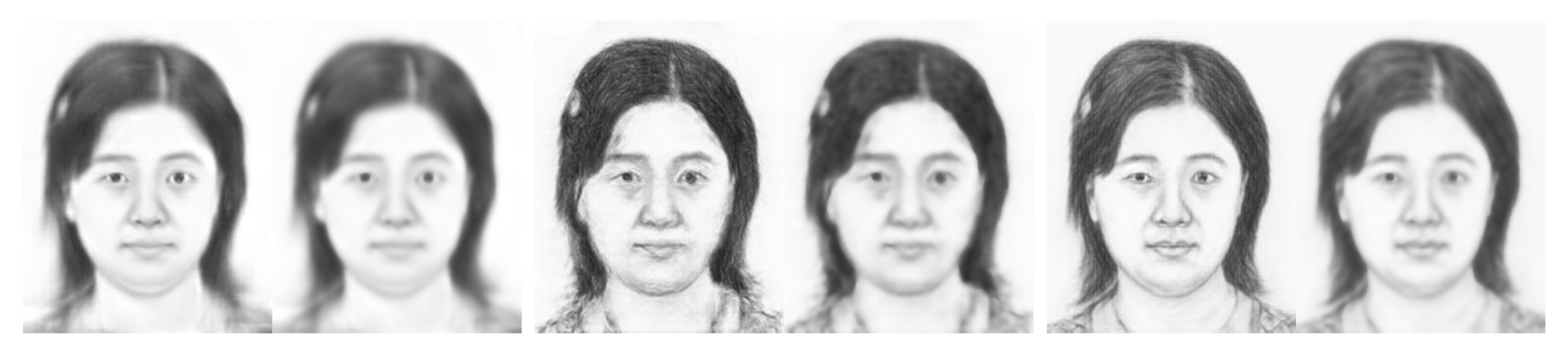}
\begin{minipage}{.32\linewidth}
\centering
(b) RSLCR \\ SSIM: 0.5970/0.5903. \\   FSIM: 0.7488/0.7362.
\end{minipage}
\begin{minipage}{.32\linewidth}
\centering
(c) Pix2Pix-GAN \\SSIM: 0.5648/0.5953. \\   FSIM: 0.7559/0.7506.
\end{minipage}
\begin{minipage}{.32\linewidth}
\centering
(d) Ours. \\ SSIM: 0.5814/0.6055. \\   FSIM: 0.7692/0.7557.
\end{minipage}

   \caption{SSIM and FSIM scores of some generated sketches (left) and their smoothed counterparts (right).}
   \label{fig:blur-example}
\end{figure}

\begin{figure}[htbp]
  \subfloat[SSIM Score on CUFS]{\includegraphics[width=0.49\linewidth]{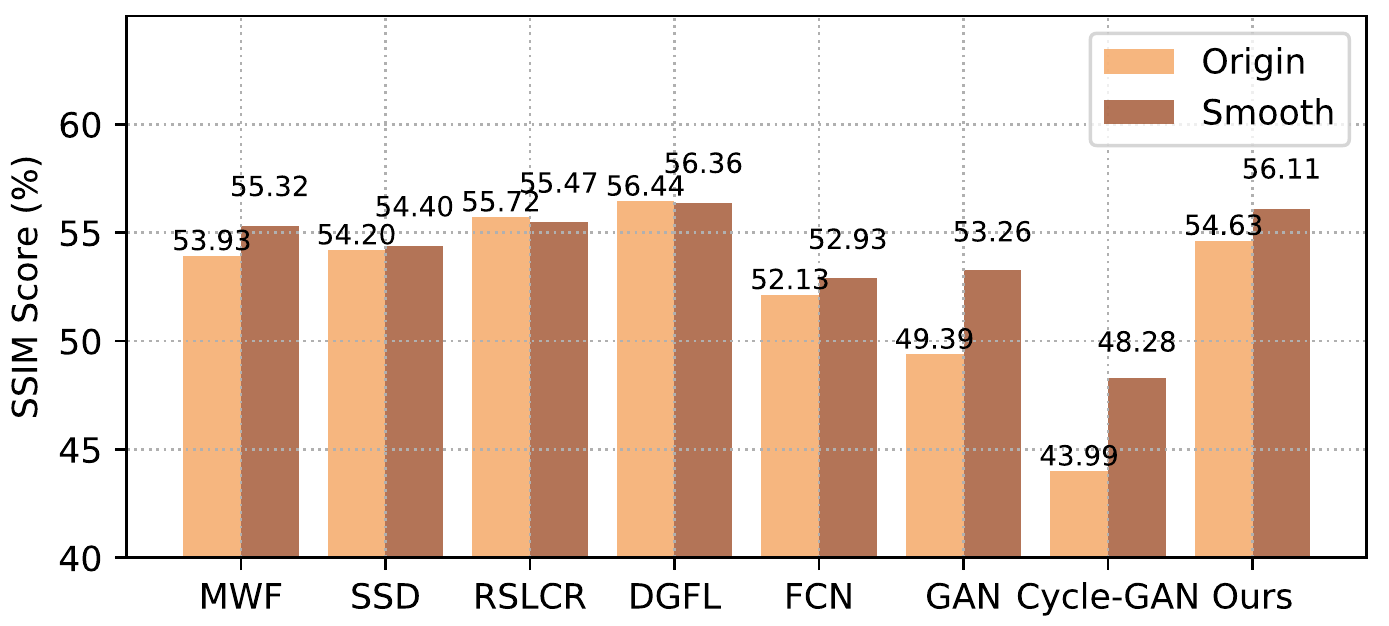}}
  \subfloat[SSIM Score on CUFSF]{\includegraphics[width=0.49\linewidth]{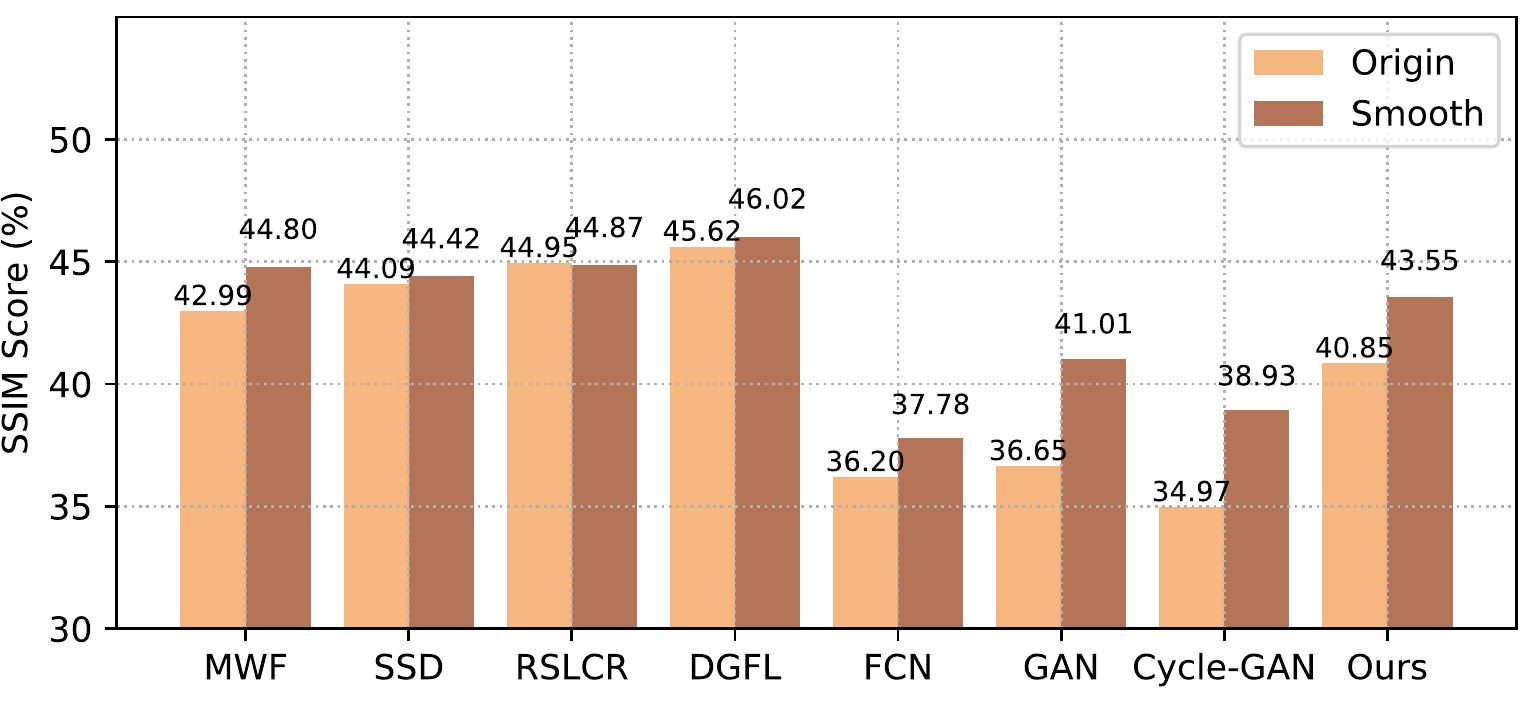}} \\
  \subfloat[FSIM Score on CUFS]{\includegraphics[width=0.49\linewidth]{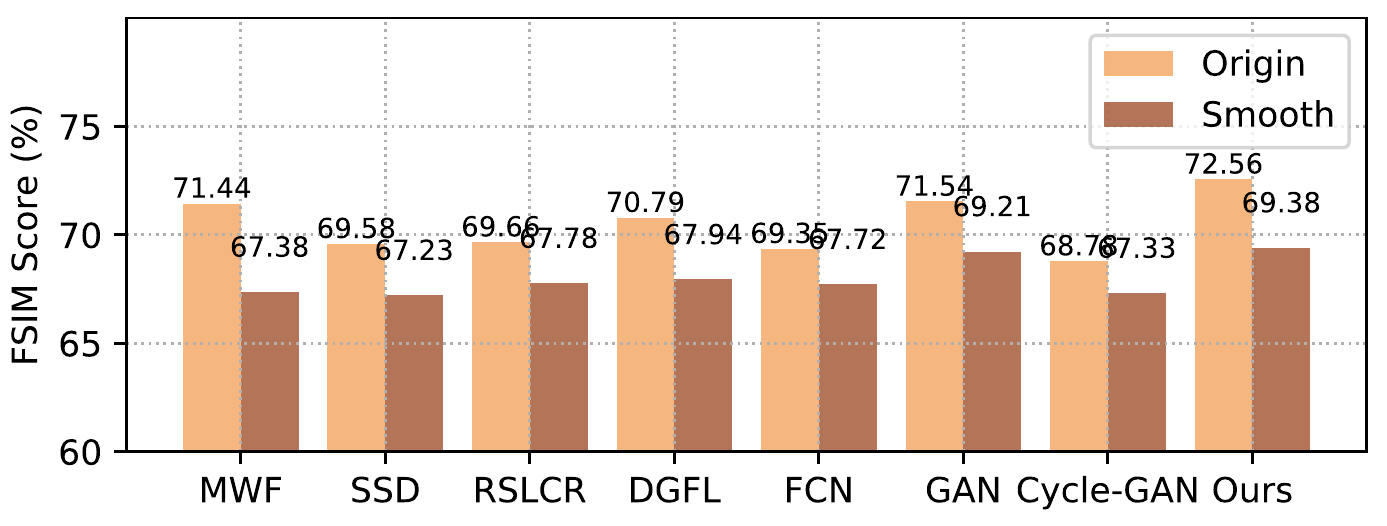}}
  \subfloat[FSIM Score on CUFSF]{\includegraphics[width=0.49\linewidth]{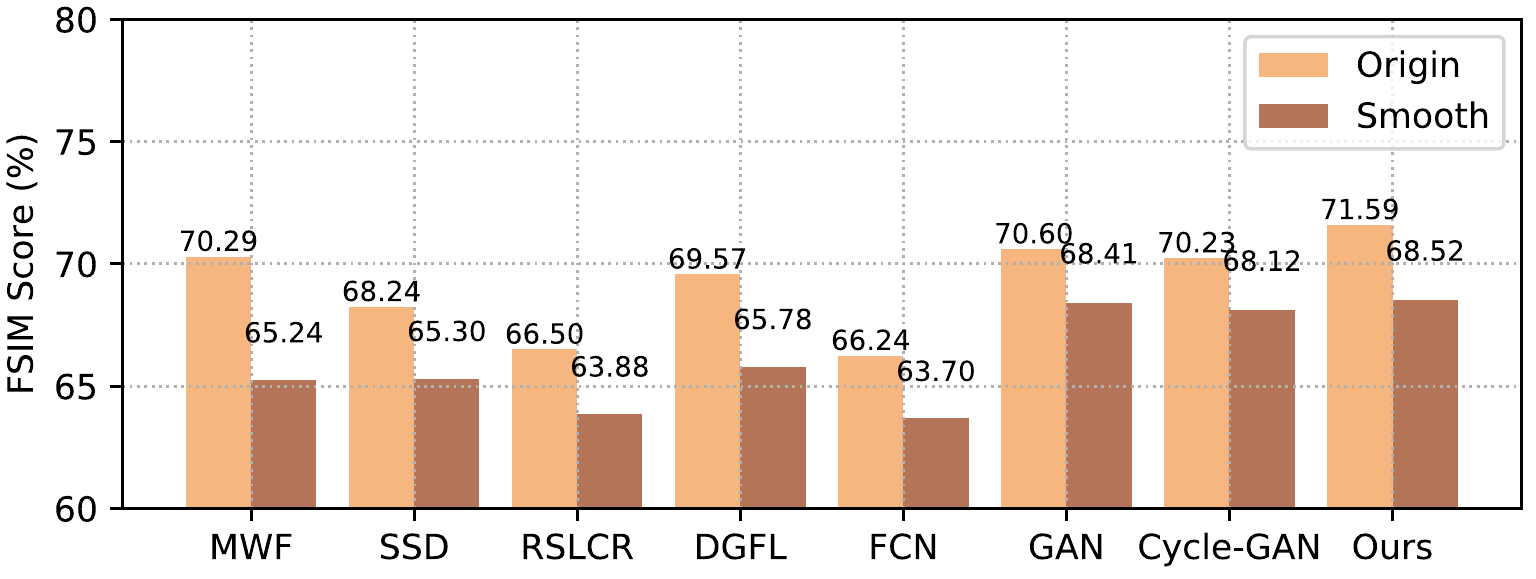}}
   \caption{Average SSIM and FSIM scores of the sketches generated by different methods on CUFS and CUFSF. The proposed method achieves state-of-the-art FSIM score on both datasets.}
   \label{fig:cufs-score}
\end{figure}

For datasets with ground truth sketches (e.g., CUFS and CUFSF), previous work \cite{wangrslcr,Gao2017cagan,ijcai2017-500} typically used structural similarity (SSIM)~\cite{karacan2013structure} as an image quality assessment metric to measure the similarity between a generated sketch and the ground truth sketch. However, many researchers (e.g., in super resolution~\cite{ledig2016photo} and face sketch synthesis~\cite{WANG2017,Wang2017psman}) pointed out that SSIM is not always consistent with the perceptual quality. One main reason is that SSIM favors slightly blurry images when the images contain rich textures. To demonstrate this, we show some sketches generated using different methods together with their SSIM scores in Fig.~\ref{fig:blur-example}. It can be seen that sketch generated by RSLCR is smoother than those by Pix2Pix-GAN and our model, but have higher SSIM scores. We applied a bilateral filter to smooth all the sketches. It can be observed that the SSIM scores of the sketch generated by RSLCR remain roughly the same after smoothing, whereas those of the sketches generated by Pix2Pix-GAN and our model improve by more than $1.5\%$.
In Fig.~\ref{fig:cufs-score}(a), we show the average SSIM scores of the sketches generated by different methods on CUFS, together with the average SSIM scores of their smoothed counterparts. As expected, the average SSIM scores of most of the methods improve after smoothing, same for a few exemplar based methods which produce over-smoothed sketches. The average SSIM score of our smoothed results is comparable to that of the state-of-the-art method. In Fig.~\ref{fig:cufs-score}(b), we show the corresponding results on CUFSF. Similar conclusions can be drawn.

\captionsetup{position=bottom}

Due to the drawback of SSIM, we use feature similarity (FSIM)~\cite{zhang2011fsim} as our image quality assessment metric. FSIM is better at evaluating detailed textures compared with SSIM. It can be observed from Fig.~\ref{fig:blur-example} that the FSIM scores of the sketches decrease after smoothing. The average FSIM scores of the sketches generated by different methods on CUFS and CUFSF are shown in Fig.~\ref{fig:cufs-score}(c) and Fig.~\ref{fig:cufs-score}(d) respectively. It can be seen that our method achieves the state-of-the-art in terms of FSIM score on both CUFS and CUFSF.

\subsubsection{Face Sketch Recognition}

\begin{figure}[htbp]
\begin{center}
   \subfloat[bottom][NLDA score on CUFS]{\includegraphics[width=0.4\linewidth]{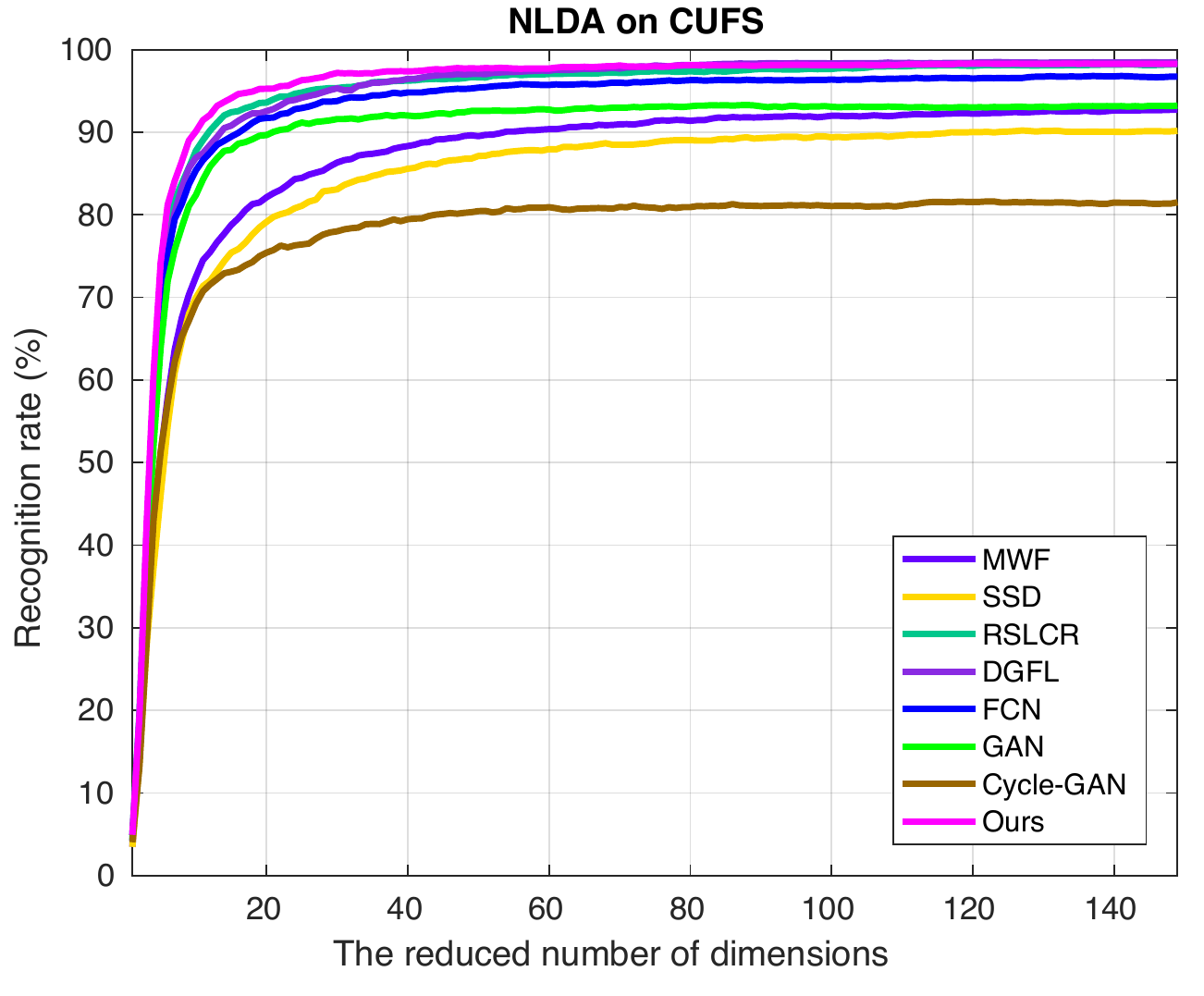}}
   \subfloat[NLDA score on CUFSF]{\includegraphics[width=0.4\linewidth]{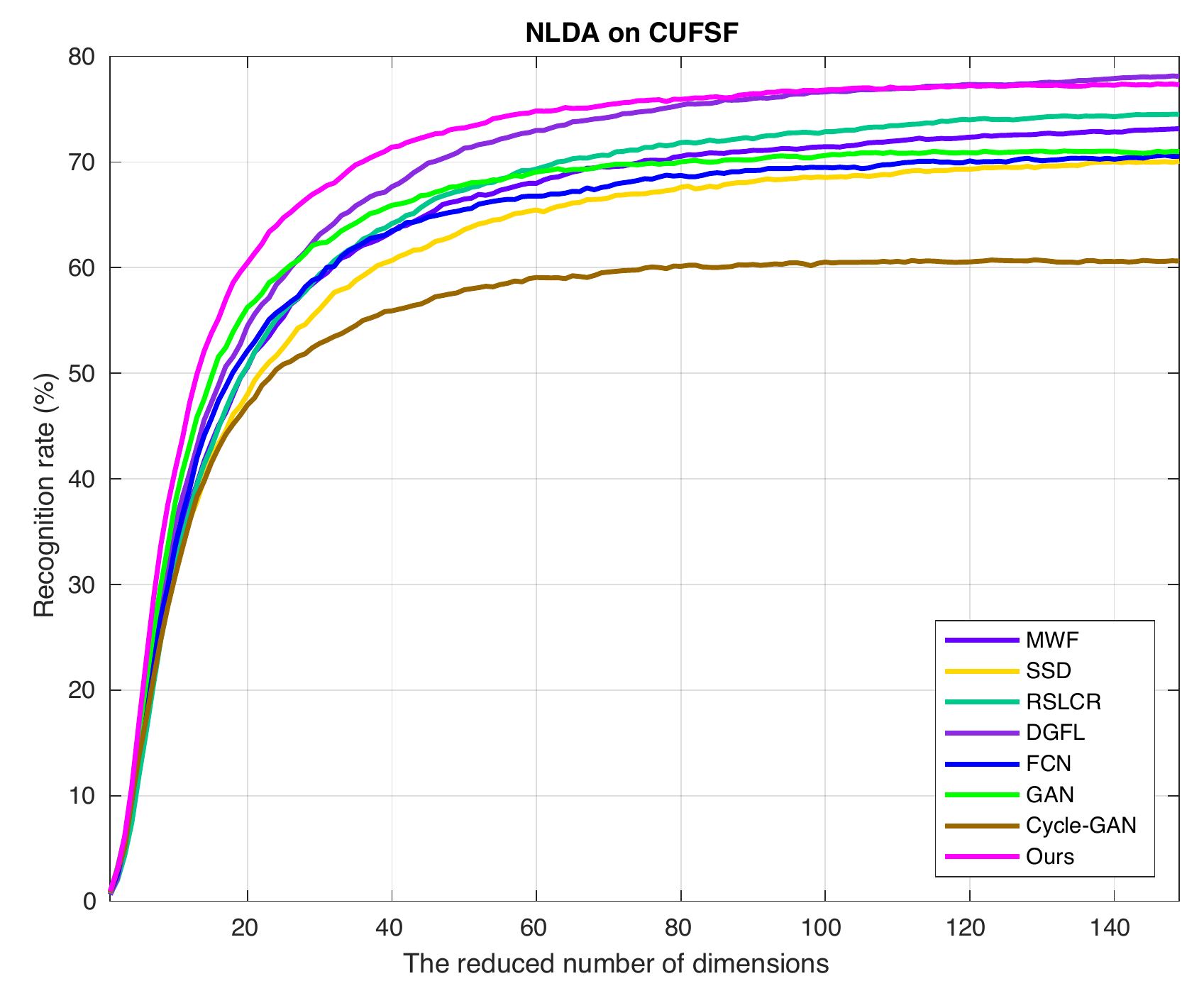}}
\end{center}
   \caption{Face recognition rate against feature dimensions on CUFS and CUFSF.}
   \label{fig:nlda-score}
\end{figure}

Sketch recognition is an important application of face sketch synthesis. We follow the same practice of Wang~\etal~\cite{wangrslcr} and employ the null-space linear discriminant analysis (NLDA)~\cite{chen2000new} to perform the recognition experiments. Fig.~\ref{fig:nlda-score} shows the recognition accuracy of different methods on the two datasets. Our method achieves the best result when the dimension of the reduced eigenspace is less than 100, and achieves a competitive result to the state-of-the-art method~\cite{ijcai2017-500} when the dimension is above 100.

\subsubsection{Comparison with PS$^2$-MAN and stack-CA-GAN} To further demonstrate the effectiveness of the proposed method, we compare it with two latest GAN methods, namely PS$^2$-MAN~\cite{Wang2017psman} and stack-CA-GAN~\cite{Gao2017cagan}, which are specially designed for sketch synthesis. As shown in Table \ref{fig:ps-ca-gan}, our method achieves the best performance on almost all datasets, except for the SSIM score in CUFSF. However, we obtain a better performance on NLDA which indicates that our model can better preserve the identify information. Note that both of these GAN methods use extra information to train their network, i.e., multi-scale supervision (PS$^2$-MAN) and parsing map (stack-CA-GAN). Compared with them, our perceptual loss can not only avoid producing artifacts but also help to improve the generalization of the network.

\begin{table}[htbp]
  \centering
  \renewcommand{\arraystretch}{1}
  \captionof{table}{Quantitative comparison with PS$^2$-MAN and stack-CA-GAN. Results are taken from their original papers.}
  \label{fig:ps-ca-gan}
  \begin{tabular}{c|c|c|c|c|c|c}
    \hline
    \multirow{2}{*}{} & \multicolumn{2}{c|}{CUHK} & \multicolumn{2}{c|}{CUFS} & \multicolumn{2}{c}{CUFSF} \\ \cline{2-7}
    & SSIM      & FSIM    & SSIM  & NLDA  & SSIM  & NLDA \\ \hline
    PS$^2$-MAN  & 0.6156  & 0.7361&  \multicolumn{2}{c|}{---} & \multicolumn{2}{c}{---} \\ \hline
    stack-CA-GAN & \multicolumn{2}{c|}{---} & 0.5266  & 96.04 & \textbf{0.4106} & 77.31 \\ \hline
    Ours       & \textbf{0.6328}   & \textbf{0.7423}& \textbf{0.5463} & \textbf{98.22} & 0.4085 & \textbf{78.04}  \\ \hline
    \end{tabular}
\end{table}

\section{Sketch Synthesis in the Wild} \label{sec:wild}
There are two challenges for sketch synthesis in the wild. The first challenge is how to deal with real photos captured under uncontrolled environments with varying pose and lighting, and cluttered backgrounds. The second is the computation time. Our method tackles the first challenge by introducing more training photos through the construction of pseudo sketch features. Regarding computation time, our CNN based model can generate a sketch in a single feed forward pass which takes about 7ms on a GPU for a $250\times200$ photo. We compared our method with five other methods , including SSD\footnote{\url{http://www.cs.cityu.edu.hk/~yibisong/eccv14/index.html}}, FCN, Pix2Pix-GAN\footnote{\url{https://github.com/phillipi/pix2pix}}, Cycle-GAN\footnote{\url{https://github.com/junyanz/pytorch-CycleGAN-and-pix2pix}}, and Fast-RSLCR\footnote{\url{http://www.ihitworld.com/RSLCR.html}}. In this experiment, we trained our model using CUFS as the reference set and all the training photos from the CUFS, CUFSF and VGG-Face10 as the training set. Since there are no ground truth sketches for the test photos, we carried out a mean opinion score (MOS) test to quantitatively evaluate the results.

\subsection{Qualitative Comparison}

\begin{figure*}[!ht]
\centering
\includegraphics[width=0.8\linewidth]{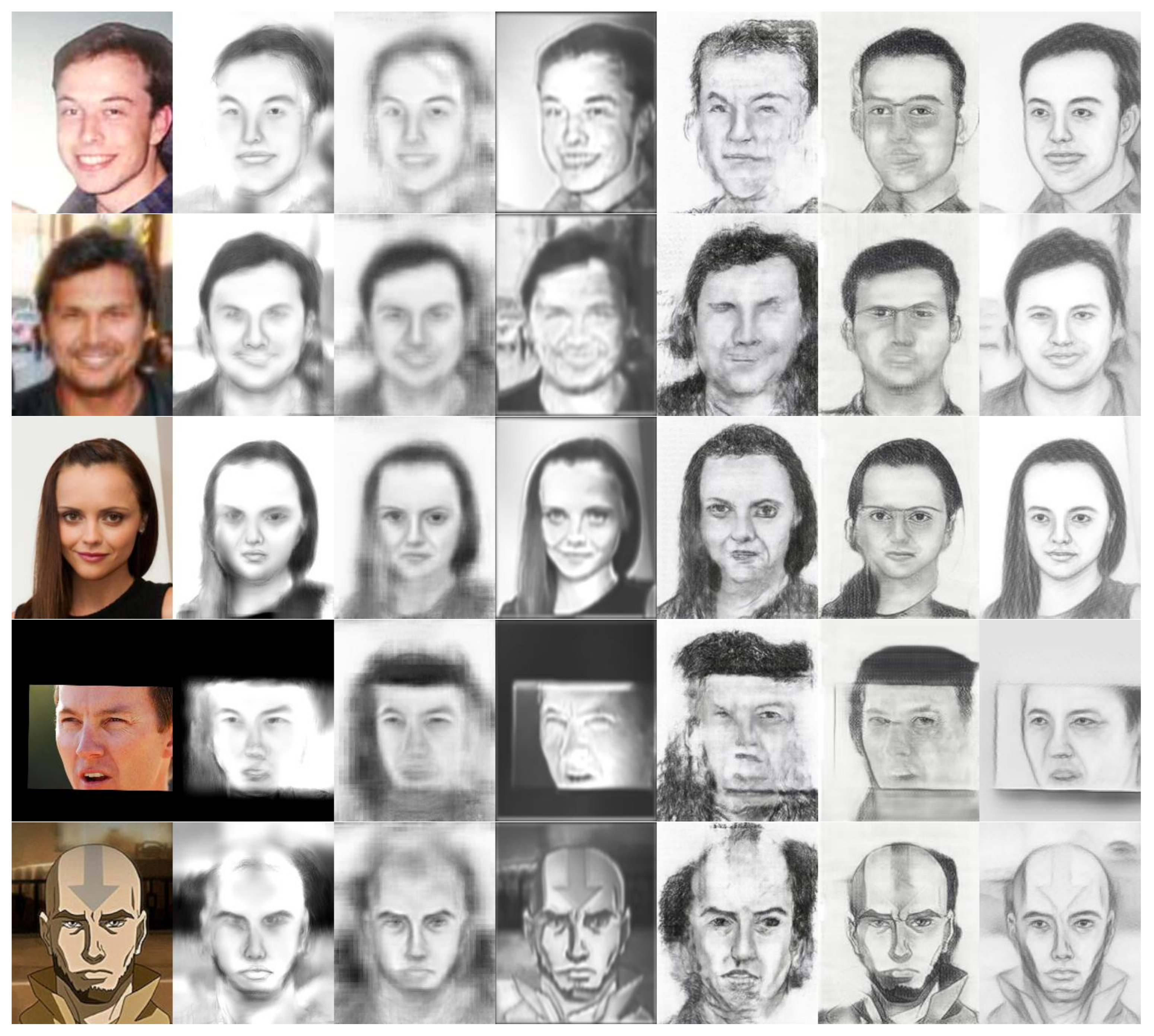}
\begin{minipage}[t]{1.\textwidth}
\centering
\tiny
\begin{tabular}{*{7}{C{1.3cm}}}
(a) Photo & (b) SSD & (c) Fast-RSLCR & (d) FCN & (e) Pix2Pix-GAN & (f) Cycle-GAN & (g) Ours
\end{tabular}
\end{minipage}
   \caption{Qualitative comparison of different methods for images in the wild. Benefit from the additional training photos, the proposed method can deal with various photos.}
   \label{fig:result-wild}
\end{figure*}

As photos in the wild are captured under uncontrolled environments, their appearance may vary largely. Fig.~\ref{fig:result-wild} shows some photos sampled from our VGG-Face test dataset and the sketches generated by different methods. It can be observed that these photos may show very different lightings, poses, image resolutions, and hair styles. Besides, some photos may be incomplete and people may also use a cartoon as their photos for entertainment (see the last row of Fig.~\ref{fig:result-wild}). It is therefore very difficult, if not impossible, for a method which only learns from a small set of photo-sketch pairs to generate sketches for photos in the wild. Among the results of other methods, exemplar based methods (see Fig.~\ref{fig:result-wild}(b)(c)) fail to deal with pose changes and different hair styles. FCN produces sketches (see Fig.~\ref{fig:result-wild}(d)) that can roughly preserve the contour of the face but lose important facial components (e.g., nose and eyes). Although GANs can generate some sketch like textures, none of them can well preserve the contents. The face shapes are distorted and the key facial parts are lost. It can be seen from Fig.~\ref{fig:result-wild}(g) that our model can handle photos in the wild well and generate pleasant results.

\subsection{Effectiveness of Additional Training Photos}
\begin{figure}[htbp]
\begin{minipage}{0.49\linewidth}
\includegraphics[width=1.\linewidth]{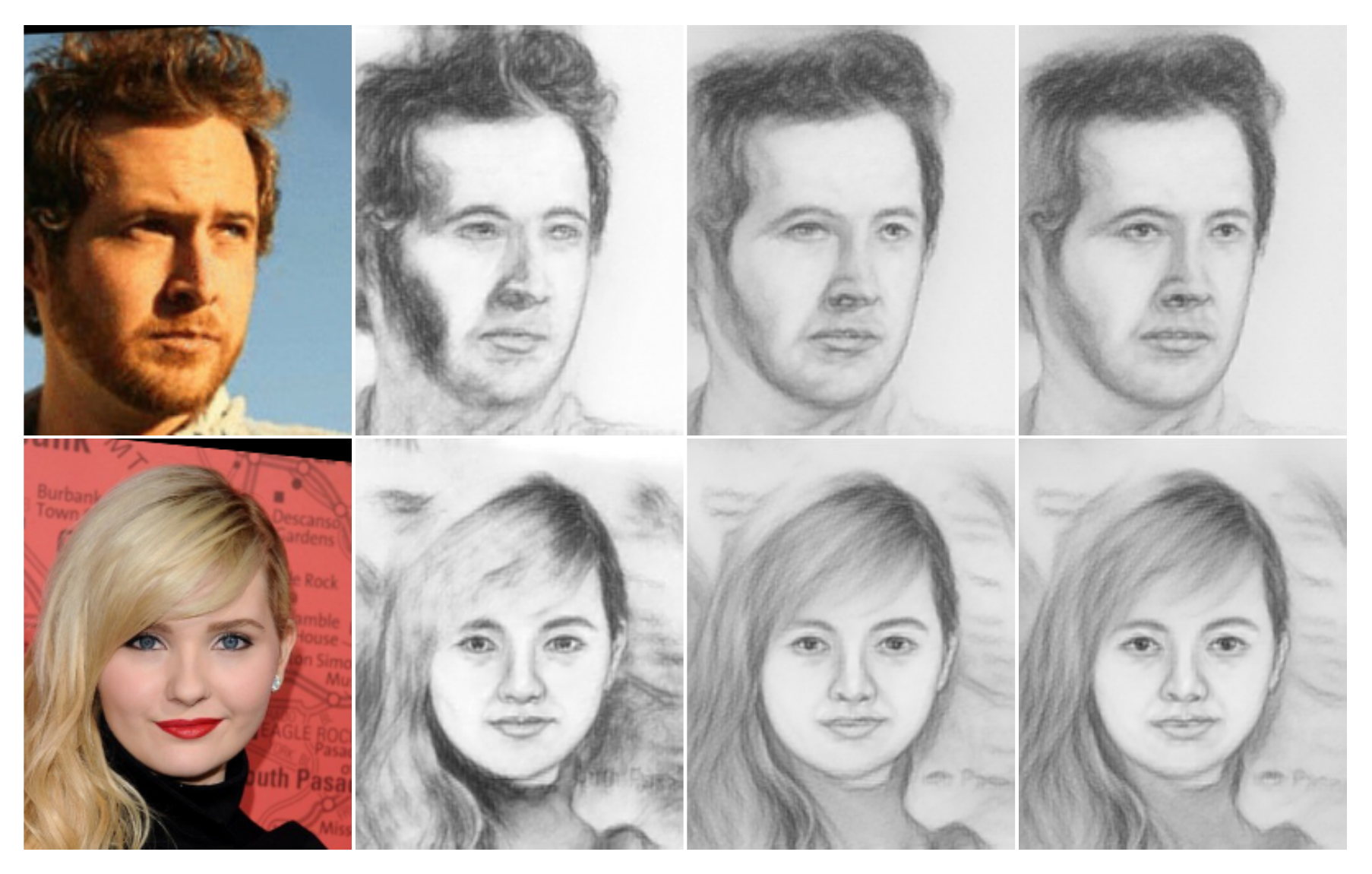}
\begin{minipage}{0.24\linewidth}
  \centering
  \tiny{(a) Photo}
\end{minipage}
\begin{minipage}{0.24\linewidth}
  \centering
  \tiny{(b) VGGFace00}
\end{minipage}
\begin{minipage}{0.24\linewidth}
  \centering
  \tiny{(c) VGGFace05}
\end{minipage}
\begin{minipage}{0.24\linewidth}
  \centering
  \tiny{(d) VGGFace10}
\end{minipage}
   \captionof{figure}{Effectiveness of additional training photos. The results improve a lot when more and more photos are added to the training set.}
   \label{fig:addition-data}
\end{minipage}
\begin{minipage}{0.45\linewidth}
\centering
   \includegraphics[width=1.\linewidth]{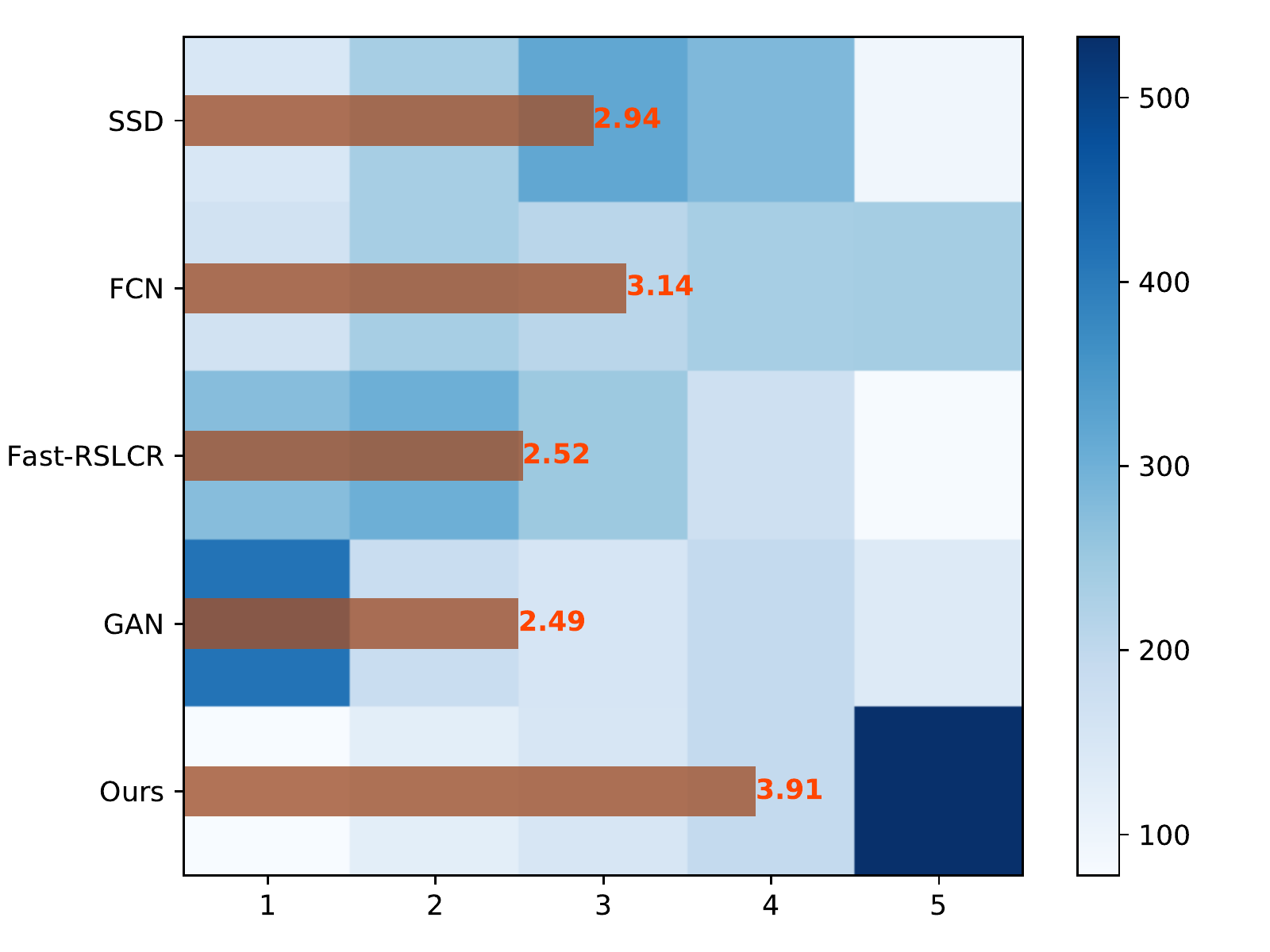}
   \captionof{figure}{Results of MOS test on the quality of sketches generated by SSD, FCN, Fast-RSLCR, Pix2Pix-GAN and our model on photos in the wild.}
   \label{fig:mos-test}
\end{minipage}
\end{figure}

Introducing more training photos from VGG-Face dataset is the key to improve the generalization ability of our model. As demonstrated in Fig.~\ref{fig:addition-data}, the model trained without additional photos from VGG-Face has difficulty in handling uncontrolled lightings and different hair colors (see Fig.~\ref{fig:addition-data}(b)). As we add more photos to the training set, the results improve significantly (see the eyes and hair in Fig.~\ref{fig:addition-data}).

\subsection{Mean Opinion Score Test}
Since there are no ground truth sketches for the photos in the wild, we performed a MOS test to assess the perceptual quality of the sketches generated by different methods. Specifically, we randomly selected 30 photos from the VGG test set, and then generated the sketches for these photos using SSD, FCN, Fast-RSLCR, Pix2Pix-GAN and our method respectively. Given the example photo-sketch pairs from public benchmarks as reference, 108 raters were asked to rank 10 groups of randomly selected sketches synthesized by the five different methods. 
We assigned a score of 1-5 to the sketches based on their rankings (5 being the best). The results are presented in Fig.~\ref{fig:mos-test}. It can be observed that the MOS of our results significantly outperforms that of the other methods. This demonstrates the superiority of our method on photos in the wild.

\section{Conclusion}
In this paper, we propose a semi-supervised learning framework for face sketch synthesis in the wild. We design a residual network with skip connections to transfer photos to sketches. Instead of supervising our network using ground truth sketches, we construct a novel pseudo sketch feature representation for each input photo based on feature space patch matching with a small reference set of photo-sketch pairs. This allows us to train our model using a large face photo dataset (without ground truth sketches) with the help of a small reference set of photo-sketch pairs. Training with a large face photo dataset enables our model to generalize better to photos in the wild. Experiments show that our method can produce sketches comparable to those produced by other state-of-the-art methods on four public benchmarks (in terms of SSIM and FSIM), and outperforms them on photos in the wild.

\section{Acknowledgment}

We thank Nannan Wang, Hao Zhou and Yibing Song for providing their codes and data. We also gratefully acknowledge the support of NVIDIA Corporation with the donation of the Titan X Pascal GPU used for this research.

\bibliographystyle{splncs}
\bibliography{egbib}

\end{document}